\documentclass[11pt,letterpaper]{article}
\usepackage{acl2012}
\usepackage{times}
\usepackage{latexsym}
\usepackage{amsmath}
\usepackage{multirow}
\usepackage[hyphens]{url}

\usepackage{arydshln}
\makeatletter
\newcommand{\@BIBLABEL}{\@emptybiblabel}
\newcommand{\@emptybiblabel}[1]{}
\makeatother
\usepackage[hidelinks]{hyperref}

\title{Learning to Understand Phrases by Embedding the Dictionary}

\author{Felix Hill\\
  Computer Laboratory \\
  University of Cambridge \\
  {\tt felix.hill@cl.cam.ac.uk} \\
  \And
  Kyunghyun Cho\thanks{
      ~~
      Work mainly done at the University of Montreal.
  }\\
Courant Institute of Mathematical Sciences \\
and Centre for Data Science \\
New York University \\ 
  {\tt kyunghyun.cho@nyu.edu} \\
  \AND
Anna Korhonen \\
Department of Theoretical and Applied Linguistics \\
University of Cambridge \\
{\tt alk23@cam.ac.uk} \\
  \And
Yoshua Bengio \\
CIFAR Senior Fellow \\
Universit\'{e} de Montr\'{e}al \\
{\tt yoshua.bengio@umontreal.ca }
}

\date{}

\begin{document}

\maketitle

\begin{abstract}

Distributional models that learn rich semantic word representations are a success story of recent NLP research. However, developing models that learn useful representations of phrases and sentences has proved far harder. We propose using the definitions found in everyday dictionaries as a means of bridging this gap between lexical and phrasal semantics. Neural language embedding models can be effectively trained to map dictionary definitions (phrases) to (lexical) representations of the words defined by those definitions. We present two applications of these architectures: \emph{reverse dictionaries} that return the name of a concept given a definition or description and general-knowledge crossword question answerers. On both tasks, neural language embedding models trained on definitions from a handful of freely-available lexical resources perform as well or better than existing commercial systems that rely on significant task-specific engineering. The results highlight the effectiveness of both neural embedding architectures and definition-based training for developing models that understand phrases and sentences. 

\end{abstract}

\section{Introduction}

Much recent research in computational semantics has focussed on learning representations of arbitrary-length phrases and sentences. This task is challenging partly because there is no obvious gold standard of phrasal representation that could be used in training and evaluation. Consequently, it is difficult to design approaches that could learn from such a gold standard, and also hard to evaluate or compare different models.

In this work, we use dictionary definitions to address this issue. The composed meaning of the words in a dictionary definition (\emph{a tall, long-necked, spotted ruminant of Africa}) should correspond to the meaning of the word they define (\emph{giraffe}). This bridge between lexical and phrasal semantics is useful because high quality vector representations of single words can be used as a target when learning to combine the words into a coherent phrasal representation.
 
This approach still requires a model capable of learning to map between arbitrary-length phrases and fixed-length continuous-valued word vectors. For this purpose we experiment with two broad classes of neural language models (NLMs): Recurrent Neural Networks (RNNs), which naturally encode the order of input words, and simpler (feedforward) bag-of-words (BOW) embedding models. Prior to training these NLMs, we learn target lexical representations by training the Word2Vec software~\cite{mikolov2013distributed} on billions of words of raw text. 

We demonstrate the usefulness of our approach by building and releasing two applications. The first is a \emph{reverse dictionary} or \emph{concept finder}: a system that returns words based on user descriptions or definitions~\cite{zock2004word}. Reverse dictionaries are used by copywriters, novelists, translators and other professional writers to find words for notions or ideas that might be on the tip of their tongue. For instance, a travel-writer might look to enhance her prose by searching for examples of a \emph{country that people associate with warm weather} or \emph{an activity that is mentally or physically demanding}. We show that an NLM-based reverse dictionary trained on only a handful of dictionaries identifies novel definitions and concept descriptions comparably or better than commercial systems, which rely on significant task-specific engineering and access to much more dictionary data. Moreover, by exploiting models that learn bilingual word representations~\cite{307754,klementiev2012inducing,hermann2013multilingual,gouws2014bilbowa}, we show that the NLM approach can be easily extended to produce a potentially useful cross-lingual reverse dictionary.

The second application of our models is as a general-knowledge crossword question answerer. When trained on both dictionary definitions and the opening sentences of Wikipedia articles, NLMs produce plausible answers to (non-cryptic) crossword clues, even those that apparently require detailed world knowledge. Both BOW and RNN models can outperform bespoke commercial crossword solvers, particularly when clues contain a greater number of words. Qualitative analysis reveals that NLMs can learn to relate concepts that are not directly connected in the training data and can thus generalise well to unseen input. To facilitate further research, all of our code, training and evaluation sets (together with a system demo) are published online with this paper.\footnote{
    \url{https://www.cl.cam.ac.uk/~fh295/}
}

\section{Neural Language Model Architectures}

The first model we apply to the dictionary-based learning task is a recurrent neural network (RNN). RNNs operate on variable-length sequences of inputs; in our case, natural language definitions, descriptions or sentences. RNNs (with LSTMs) have achieved state-of-the-art performance in language modelling~\cite{mikolov2010recurrent}, image caption generation~\cite{kiros2014unifying} and approach state-of-the-art performance in machine translation~\cite{bahdanau2014neural}. 

During training, the input to the RNN is a dictionary definition or sentence from an encyclopedia. The objective of the model is to map these defining phrases or sentences to an embedding of the word that the definition defines. The target word embeddings are learned independently of the RNN weights, using the Word2Vec software~\cite{mikolov2013distributed}.   

The set of all words in the training data constitutes the vocabulary of the RNN. For each word in this vocabulary we randomly initialise a real-valued vector (input embedding) of model parameters. The RNN `reads' the first word in the input by applying a non-linear projection of its embedding \(v_1\) parameterised by input weight matrix \(W\) and \(b\), a vector of biases.
\begin{align*}
A_1 = \phi( Wv_1 + b) 
\end{align*}
yielding the first internal activation state \(A_1\). In our implementation, we
use \(\phi(x) = \tanh(x)\), though in theory \(\phi\) can be any differentiable
non-linear function. Subsequent internal activations (after time-step \(t\))
are computed by projecting the embedding of the \(t^{th}\) word and using this
information to `update' the internal activation state. 
\begin{align*}
A_t =  \phi( UA_{t-1} + Wv_t + b ). 
\end{align*}

As such, the values of the final internal activation state units \(A_N\) are a
weighted function of all input word embeddings, and constitute a `summary' of
the information in the sentence.

\subsection{Long Short Term Memory}

A known limitation when training RNNs to read language using gradient descent is that the error signal (gradient) on the training examples either vanishes or explodes as the number of time steps (sentence length) increases \cite{bengio1994learning}. Consequently, after reading longer sentences the final internal activation \(A_N\) typically retains useful information about the most recently read (sentence-final) words, but can neglect important information near the start of the input sentence. LSTMs  \cite{hochreiter1997long} were designed to mitigate this long-term dependency problem. 

At each time step \(t\), in place of the single internal layer of units \(A\),
the LSTM RNN computes six internal layers \(i^w , g^i, g^f , g^o, h\) and
\(m\). The first, \(g^w\), represents the core information passed to the LSTM
unit by the latest input word at \(t\). It is computed as a simple linear
projection of the input embedding \(v_t\) (by input weights \(W_w\)) and the
\emph{output state} of the LSTM at the previous time step \(h_{t-1}\) (by
update weights \(U_w\)):
\begin{align*}
i_t^w = W_w v_t + U_wh_{t-1} +  b_w
\end{align*}

The layers \(g^i, g^f \) and \(g^o \) are computed as weighted sigmoid
functions of the input embeddings, again parameterised by layer-specific weight
matrices \(W\) and \(U\):
\begin{align*}
g_t^s = \frac{1}{1+\exp(-(W_s v_t + U_sh_{t-1} + b_s  ))}
\end{align*}
where \(s\) stands for one of \(i,f\) or \(o\). These vectors take
values on \([0,1]\) and are often referred to as \emph{gating activations}.
Finally, the \emph{internal memory state}, \(m_t\) and new output state
\(h_t\), of the LSTM at \(t\) are computed as
\begin{align*}
    m_t =& i_t^w \odot g_t^i + m_{t-1} \odot g_t^f \\
    h_t =& g_t^o \odot \phi(m_t), 
\end{align*}
where \(\odot\) indicates elementwise vector multiplication and \(\phi\) is, as
before, some non-linear function (we use \(tanh\)). Thus, \(g^i\) determines to
what extent the new \emph{input} word is considered at each time step, \(g^f\)
determines to what extent the existing state of the internal memory is retained
or \emph{forgotten} in computing the new internal memory, and \(g^o\)
determines how much this memory is considered when computing the output state
at \(t\). 

The sentence-final memory state of the LSTM, \(m_N\), a `summary' of all the
information in the sentence, is then projected via an extra non-linear
projection (parameterised by a further weight matrix) to a target embedding
space. This  layer enables the target (defined) word embedding space to take a
different dimension to the activation layers of the RNN, and in principle
enables a more complex definition-reading function to be learned. 

\subsection{Bag-of-Words NLMs}

We implement a simpler linear bag-of-words (BOW) architecture for encoding the
definition phrases. As with the RNN, this architecture learns an embedding
\(v_i\) for each word in the model vocabulary, together with a single matrix of
input projection weights \(W\). The BOW model simply maps an input definition
with word embeddings \(v_1 \dots v_n\) to the sum of the projected embeddings
\(\sum_{i=1}^n Wv_i \). This model can also be considered a special case of an
RNN in which the update function \(U\) and nonlinearity \(\phi\) are both the
identity, so that `reading' the next word in the input phrase updates the
current representation more simply:
\begin{align*}
A_t =  A_{t-1} + Wv_t.
\end{align*}

\subsection{Pre-trained Input Representations}

We experiment with variants of these models in which the input definition embeddings are pre-learned and fixed (rather than randomly-initialised and updated) during training. There are several potential advantages to taking this approach. First, the word embeddings are trained on massive corpora and may therefore introduce additional linguistic or conceptual knowledge to the models. Second, at test time, the models will have a larger effective vocabulary, since the pre-trained word embeddings typically span a larger vocabulary than the union of all dictionary definitions used to train the model. Finally, the models will then map to and from the same space of embeddings (the embedding space will be closed under the operation of the model), so conceivably could be more easily applied as a general-purpose `composition engine'.

\subsection{Training Objective}

We train all neural language models \(M\) to map the input definition phrase \(s_c\) defining word \(c\) to a location close to the the pre-trained embedding \(v_c\) of \(c\). We experiment with two different cost functions for the word-phrase pair \((c,s_c)\) from the training data. The first is simply the cosine distance between \(M(s_c)\) and \(v_c\). The second is the rank loss 
\[
\max(0, m - cos(M(s_c),v_c)-\cos(M(s_c),v_r))
\] 
where \(v_r\) is the embedding of a randomly-selected word from the vocabulary other than \(c\). This loss function was used for language models, for example, in~\cite{huang2012improving}. In all experiments we apply a margin \(m = 0.1\), which has been shown to work well on word-retrieval tasks~\cite{bordes2015large}. 

\subsection{Implementation Details}
Since training on the dictionary data took 6-10 hours, we did not conduct a hyper-parameter search on any validation sets over the space of possible model configurations such as embedding dimension, or size of hidden layers. Instead, we chose these parameters to be as standard as possible based on previous research. For fair comparison, any aspects of model design that are not specific to a particular class of model were kept constant across experiments. 

The pre-trained word embeddings used in all of our models (either as input or target) were learned by a continuous bag-of-words (CBOW) model using the Word2Vec software on approximately 8 billion words of running text.\footnote{The Word2Vec embedding models are well known; further details can be found at~\url{https://code.google.com/p/word2vec/} The training data for this pre-training was compiled from various online text sources using the script \emph{demo-train-big-model-v1.sh} from the same page.} When training such models on massive corpora, a large embedding length of up to 700 have been shown to yield best performance (see e.g.~\cite{faruqui2014retrofitting}). The pre-trained embeddings used in our models were of length 500, as a compromise between quality and memory constraints.

In cases where the word embeddings are learned during training on the dictionary objective, we make these embeddings shorter (256), since they must be learned from much less language data. In the RNN models, and at each time step each of the four LSTM RNN internal layers (gating and activation states) had length 512 -- another standard choice (see e.g.~\cite{cho2014learning}). The final hidden state was mapped linearly to length 500, the dimension of the target embedding. In the BOW models, the projection matrix projects input embeddings (either learned, of length 256, or pre-trained, of length 500) to length 500 for summing. 

All models were implemented with Theano~\cite{bergstra+al:2010-scipy} and trained with minibatch SGD on GPUs. The batch size was fixed at 16 and the learning rate was controlled by \emph{adadelta}~\cite{zeiler2012adadelta}. 

\section{Reverse Dictionaries}

The most immediate application of our trained models is as a \emph{reverse dictionary} or \emph{concept finder}. It is simple to look up a definition in a dictionary given a word, but professional writers often also require suitable words for a given idea, concept or definition.\footnote{See the testimony from professional writers at \url{http://www.onelook.com/?c=awards}} Reverse dictionaries satisfy this need by returning candidate words given a phrase, description or definition. For instance, when queried with the phrase \emph{an activity that requires strength and determination}, the OneLook.com reverse dictionary returns the concepts \emph{exercise} and \emph{work}. Our trained RNN model can perform a similar function, simply by mapping a phrase to a point in the target (Word2Vec) embedding space, and returning the words corresponding to the embeddings that are closest to that point.  

Several other academic studies have proposed reverse dictionary models. These generally rely on common techniques from information retrieval, comparing definitions in their internal database to the input query, and returning the word whose definition is `closest' to that query~\cite{bilac2003improving,bilac2004dictionary,zock2004word}. Proximity is quantified differently in each case, but is generally a function of hand-engineered features of the two sentences. For instance,~\newcite{shaw2013building} propose a method in which the candidates for a given input query are all words in the model's database whose definitions contain one or more words from the query. This candidate list is then ranked according to a query-definition similarity metric based on the hypernym and hyponym relations in WordNet, features commonly used in IR such as \emph{tf-idf} and a parser. 

There are, in addition, at least two commercial online reverse dictionary applications, whose architecture is proprietary knowledge. The first is the Dictionary.com reverse dictionary \footnote{Available at \url{http://dictionary.reference.com/reverse/}}, which retrieves candidate words from the Dictionary.com dictionary based on user definitions or descriptions. The second is {\bf OneLook.com}, whose algorithm searches 1061 indexed dictionaries, including all major freely-available online dictionaries and resources such as Wikipedia and WordNet.

\subsection{Data Collection and Training}

To compile a bank of dictionary definitions for training the model, we started with all words in the target embedding space. For each of these words, we extracted dictionary-style definitions from five electronic resources: \emph{Wordnet, The American Heritage Dictionary, The Collaborative International Dictionary of English, Wiktionary} and \emph{Webster's}. We chose these five dictionaries because they are freely-available via the WordNik API,\footnote{See \url{http://developer.wordnik.com}} but in theory any dictionary could be chosen. Most words in our training data had multiple definitions. For each word \(w\) with definitions \( \{d_1 \dots d_n\} \) we included all pairs \((w, d_1) \dots (w,d_n) \) as training examples. 

To allow models access to more factual knowledge than might be present in a dictionary (for instance, information about specific entities, places or people, we supplemented this training data with information extracted from Simple Wikipedia.~\footnote{\url{https://simple.wikipedia.org/wiki/Main_Page}} For every word in the model's target embedding space that is also the title of a Wikipedia article, we treat the sentences in the first paragraph of the article as if they were (independent) definitions of that word. When a word in Wikipedia also occurs in one (or more) of the five training dictionaries, we simply add these pseudo-definitions to the training set of definitions for the word. Combining Wikipedia and dictionaries in this way resulted in \(\approx 900,000\) word-'definition' pairs of \(\approx 100,000\) unique words. 

To explore the effect of the quantity of training data on the performance of the models, we also trained models on subsets of this data. The first subset comprised only definitions from Wordnet (approximately 150,000 definitions of 75,000 words). The second subset comprised only words in Wordnet and their \emph{first} definitions (approximately 75,000 word, definition pairs).\footnote{As with other dictionaries, the first definition in WordNet generally corresponds to the most typical or common sense of a word.}. For all variants of RNN and BOW models, however, reducing the training data in this way resulted in a clear reduction in performance on all tasks. For brevity, we therefore do not present these results in what follows.  

\begin{table*}[ht]
    \centering
{\small
\hfill{}
\begin{tabular}{r|r|ccc|ccc|ccc|}

\multicolumn{2}{c}{}& \multicolumn{6}{|c|}{\bf Dictionary definitions} \\
\multicolumn{2}{c}{\textbf{Test Set }}&\multicolumn{3}{|c|}{\textbf{Seen} (500 WN defs)}& \multicolumn{3}{|c|}{\textbf{Unseen} (500 WN defs)} & \multicolumn{3}{|c|}{\textbf{Concept descriptions} (200)} \\

\hline

\rule{0pt}{2ex} 

Unsup. & W2V add & - & - & - & 923 & .04/.16 & 163 & 339 & .07/.30 & 150    \\
models  & W2V mult &- &- & -& 1000 & .00/.00 & 10* &   1000 & .00/.00 & 27* \\
\hdashline 
\rule{0pt}{2ex} 
& OneLook & \bf 0 & \bf .89/.91 & \bf 67  & - & - & - &  \bf 18.5 &  {\bf .38}/.58 & 153    \\
\hdashline 
\rule{0pt}{2ex} 
 & RNN cosine & 12 & .48/.73 & 103 &  22 & .41/.70 & 116 & 69 & .28/.54 & 157 \\
 & RNN w2v cosine & 19 & .44/.70 & 111 & 19 & .44/.69 & 126 & 26 & {\bf .38}/.66 & 111  \\
 & RNN ranking & 18 & .45/.67 & 128 &	24 & .43/.69 & 103 & 25 & .34/.66 & 102 \\
NLMs & RNN w2v ranking & 54 & .32/.56 & 155 & 33 & .36/.65 & 137 &	30 & .33/.69 & \bf 77 \\
& BOW cosine &22 & .44/.65 & 129 & 19 & .43/.69 & 103 & 50 & .34/.60 &  99 \\
& BOW w2v cosine & 15 & .46/.71 & 124 &  \bf14 & \bf .46/ .71 &  104	 & 28 & .36/.66 &  99 \\
& BOW ranking & 17 & .45/.68 &  115 &	 22 & .42/.70 & \bf 95 &	32 & .35/.69 & 101   \\
& BOW w2v rankng & 55 & .32/.56 & 155 &	36 & .35/.66 & 138 &	38 & .33/{\bf .72} & 85 \\

\hline 

\multicolumn{11}{c}{} \\
\multicolumn{5}{c}{}& \multicolumn{6}{|c|}{\emph{median rank \hspace{5mm}   accuracy@10/100 \hspace{5mm}   rank variance} } \\

\end{tabular}}
\caption{Performance of different reverse dictionary models in different evaluation settings. *Low variance in \emph{mult} models is due to consistently poor scores, so not highlighted.}
\label{results}
\end{table*}

\subsection{Comparisons}

As a baseline, we also implemented two entirely unsupervised methods using the neural (Word2Vec) word embeddings from the target word space. In the first ({\bf W2V add}), we compose the embeddings for each word in the input query by pointwise addition, and return as candidates the nearest word embeddings to the resulting composed vector.\footnote{Since we retrieve all answers from embedding spaces by cosine similarity, addition of word embeddings is equivalent to taking the mean.} The second baseline, ({\bf W2V mult}), is identical except that the embeddings are composed by elementwise multiplication. Both methods are established ways of building phrase representations from word embeddings~\cite{mitchell2010composition}.

None of the models or evaluations from previous academic research on reverse dictionaries is publicly available, so direct comparison is not possible. However, we do compare performance with the commercial systems. The Dictionary.com system returned no candidates for over 96\% of our input definitions. We therefore conduct detailed comparison with OneLook.com, which is the first reverse dictionary tool returned by a Google search and seems to be the most popular among writers. 

\subsection{Reverse Dictionary Evaluation}

To our knowledge there are no established means of measuring reverse dictionary performance. In the only previous academic research on English reverse dictionaries that we are aware of, evaluation was conducted on 300 word-definition pairs written by lexicographers~\cite{shaw2013building}. Since these are not publicly available we developed new evaluation sets and make them freely available for future evaluations.  

The evaluation items are of three types, designed to test different properties of the models. To create the {\bf seen} evaluation, we randomly selected 500 words from the WordNet training data (seen by all models), and then randomly selected a definition for each word. Testing models on the resulting 500 word-definition pairs assesses their ability to recall or decode previously encoded information. For the {\bf unseen} evaluation, we randomly selected 500 words from WordNet and excluded all definitions of these words from the training data of all models. 

Finally, for a fair comparison with OneLook, which has both the seen and unseen pairs in its internal database, we built a new dataset of {\bf concept descriptions} that do not appear in the training data for any model. To do so, we randomly selected 200 adjectives, nouns or verbs from among the top 3000 most frequent tokens in the British National Corpus~\cite{leech1994claws4} (but outside the top 100). We then asked ten native English speakers to write a single-sentence `description' of these words. To ensure the resulting descriptions were good quality, for each description we asked two participants who did not produce that description to list any words that fitted the description (up to a maximum of three). If the target word was not produced by one of the two checkers, the original participant was asked to re-write the description until the validation was passed.\footnote{Re-writing was required in 6 of the 200 cases.} These concept descriptions, together with other evaluation sets, can be downloaded from our website for future comparisons.

\begin{table}[ht]
{\small
\emph
\hfill{}
\begin{tabular}{r|cl}
\bf Test set & \bf Word & \bf Description \\
\hline
 Dictionary &   \emph{valve} & "control consisting of a mechanical   \\
definition  & & device for controlling fluid flow" \\ 
\rule{0pt}{2ex} 
Concept &   \emph{prefer} & "when you like one thing \\
description & & more than another thing" \\
\end{tabular}
\caption{Style difference between \emph{dictionary definitions} and \emph{concept descriptions} in the evaluation.}
\label{tb:tablename}}
\end{table}

Given a test description, definition, or question, all models produce a ranking of possible word answers based on the proximity of their representations of the input phrase and all possible output words. To quantify the quality of a given ranking, we report three statistics: the \emph{median rank} of the correct answer (over the whole test set, lower better), the proportion of training cases in which the correct answer appears in the top 10/100 in this ranking (\emph{accuracy@10/100} - higher better) and the variance of the rank of the correct answer across the test set (\emph{rank variance} - lower better). 

\subsection{Results}

Table~\ref{results} shows the performance of the different models in the three evaluation settings. Of the unsupervised composition models, elementwise addition is clearly more effective than multiplication, which almost never returns the correct word as the nearest neighbour of the composition. Overall, however, the supervised models (RNN, BOW and OneLook) clearly outperform these baselines. 

The results indicate interesting differences between the NLMs and the OneLook dictionary search engine. The Seen (WN first) definitions in Table~\ref{results} occur in both the training data for the NLMs and the lookup data for the OneLook model. Clearly the OneLook algorithm is better than NLMs at retrieving already available information (returning 89\% of  correct words among the top-ten candidates on this set). However, this is likely to come at the cost of a greater memory footprint, since the model requires access to its database of dictionaries at query time.\footnote{The trained neural language models are approximately half the size of the six training dictionaries stored as plain text, so would be hundreds of times smaller than the OneLook database of 1061 dictionaries if stored this way.}

The performance of the NLM embedding models on the (unseen) concept descriptions task shows that these models can generalise well to novel, unseen queries. While the median rank for OneLook on this evaluation is lower, the NLMs retrieve the correct answer in the top ten candidates approximately as frequently, within the top 100 candidates more frequently and with lower variance in ranking over the test set. Thus, NLMs seem to generalise more `consistenly' than OneLook on this dataset, in that they generally assign a reasonably high ranking to the correct word. In contrast, as can also be verified by querying our we demo, OneLook tends to perform either very well or poorly on a given query.\footnote{We also observed that the \emph{mean} ranking for NLMs was lower than for OneLook on the concept descriptions task.}

When comparing between NLMs, perhaps the most striking observation is that the RNN models do not significantly outperform the BOW models, even though the BOW model output is invariant to changes in the order of words in the definition. Users of the online demo can verify that the BOW models recover concepts from descriptions strikingly well, even when the words in the description are permuted. This observation underlines the importance of lexical semantics in the interpretation of language by NLMs, and is consistent with some other recent work on embedding sentences \cite{iyyer2015deep}.    

It is difficult to observe clear trends in the differences between NLMs that
learn input word embeddings and those with pre-trained (Word2Vec) input
embeddings. Both types of input yield good performance in some situations and
weaker performance in others. In general, pre-training input embeddings seems
to help most on the concept descriptions, which are furthest from the training
data in terms of linguistic style. This is perhaps unsurprising, since models
that learn input embeddings from the dictionary data acquire all of their
conceptual knowledge from this data (and thus may overfit to this setting),
whereas models with pre-trained embeddings have some semantic memory acquired
from general running-text language data and other knowledge acquired from the
dictionaries.

\begin{table*}[ht]
{\small
\emph
\hfill{}
\begin{tabular}{r|ccccc|}
\bf Input & \\
\bf Description & \bf OneLook & \bf W2V add &  \bf RNN  & \bf BOW \\
\hline

\rule{0pt}{3ex} 

  "a native of  & 1:\emph{country} 2:\emph{citizen} &  1:\emph{a} 2.\emph{the}   &  1:\emph{eskimo} 2:\emph{scandinavian}   &  1:\emph{frigid} 2:\emph{cold}    \\

a cold  & 3:\emph{foreign} 4:\emph{naturalize} &   3:\emph{another} 4:\emph{of}  & 3:\emph{arctic} 4:\emph{indian}  & 3:\emph{icy} 4:\emph{russian}\\
 country" & 5:\emph{cisco} &  5:\emph{whole} &  5:\emph{siberian}  &  5:\emph{indian} \\
\rule{0pt}{3ex} 
  "a way of & 1:\emph{drag} 2:\emph{whiz} &  1:\emph{the} 2:\emph{through}   &  1:\emph{glide} 2:\emph{scooting}  &  1:\emph{flying} 2:\emph{gliding} \\

moving  & 3:\emph{aerodynamics} 4:\emph{draught} &   3:\emph{a} 4:\emph{moving}  & 3:\emph{glides} 4:\emph{gliding}  & 3:\emph{glide} 4:\emph{fly}\\
 through & 5:\emph{coefficient of drag} &  5:\emph{in} &  5:\emph{flight} &  5:\emph{scooting}\\	
 the air"        & \\

\rule{0pt}{3ex} 
  "a habit that & 1:\emph{sisterinlaw} 2:\emph{fatherinlaw} &  1:\emph{annoy} 2:\emph{your}   &  1:\emph{bossiness} 2:\emph{jealousy} &  1:\emph{infidelity} 2:\emph{bossiness}  \\

might annoy & 3:\emph{motherinlaw} 4:\emph{stepson} &   3:\emph{might} 4:\emph{that}  & 3:\emph{annoyance} 4:\emph{rudeness} & 3:\emph{foible} 4:\emph{unfaithfulness}\\
 your spouse" & 5:\emph{stepchild} &  5:\emph{either} &  5:\emph{boorishness} &  5:\emph{adulterous} \\

\end{tabular}}
\hfill{}
\caption{The top-five candidates for example queries (invented by the authors) from different reverse dictionary models. Both the RNN and BOW models are without Word2Vec input and use the cosine loss.}
\label{qual}
\end{table*}

\subsection{Qualitative Analysis}

Some example output from the various models is presented in Table~\ref{qual}. The differences illustrated here are also evident from querying the web demo. The first example shows how the NLMs (BOW and RNN) generalise beyond their training data. Four of the top five responses could be classed as appropriate in that they refer to inhabitants of cold countries. However, inspecting the WordNik training data, there is no mention of \emph{cold} or anything to do with climate in the definitions of \emph{Eskimo}, \emph{Scandinavian}, \emph{Scandinavia} etc. Therefore, the embedding models must have learned that \emph{coldness} is a characteristic of Scandinavia, Siberia, Russia, relates to Eskimos etc. via connections with other concepts that are described or defined as \emph{cold}. In contrast, the candidates produced by the OneLook and (unsupervised) W2V baseline models have nothing to do with coldness.

The second example demonstrates how the NLMs generally return candidates whose linguistic or conceptual function is appropriate to the query. For a query referring explicitly to a means, method or process, the RNN and BOW models produce verbs in different forms or an appropriate deverbal noun. In contrast, OneLook returns words of all types (\emph{aerodynamics, draught}) that are arbitrarily related to the words in the query. A similar effect  is apparent in the third example. While the candidates produced by the OneLook model are the correct part of speech (Noun), and related to the query topic, they are not semantically appropriate. The dictionary embedding models are the only ones that return a list of plausible \emph{habits}, the class of noun requested by the input.

\subsection{Cross-Lingual Reverse Dictionaries}

\begin{table*}[ht]
{\small
\emph
\hfill{}
\begin{tabular}{r|ccccc|}
\bf Input description & \bf RNN EN-FR & \bf W2V add &  \bf RNN + Google  \\
\hline
  "an emotion that you might feel & \emph{ \underline{triste}, \underline{pitoyable}} & \emph {insister, effectivement} & \emph{ sentiment, regretter} \\

after being rejected" & \emph{ \underline{r\'epugnante}, \underline{\'epouvantable}} & \emph{pourquoi, nous} &\emph{  \underline{peur, aversion} } \\
\rule{0pt}{3ex} 
"a small black flying insect that  & \emph{\underline{mouche}, canard} & \emph {attentivement, pouvions} & \emph{ voler, \underline{faucon}} \\ 
transmits disease and likes horses" &  \emph{  \underline{hirondelle}, pigeon} & \emph{pourrons, naturellement} & \emph{\underline{mouches}, volant} \\
\end{tabular}}
\hfill{}
\caption{Responses from cross-lingual reverse dictionary models to selected queries. Underlined responses are `correct' or potentially useful for a native French speaker.}
\label{cross}
\end{table*}

We now show how the RNN architecture can be easily modified to create a \emph{bilingual reverse dictionary} - a system that returns candidate words in one language given a description or definition in another. A bilingual reverse dictionary could have clear applications for translators or transcribers. Indeed, the problem of attaching appropriate words to concepts may be more common when searching for words in a  second language than in a monolingual context.  

To create the bilingual variant, we simply replace the Word2Vec target embeddings with those from a bilingual embedding space. Bilingual embedding models use bilingual corpora to learn a space of representations of the words in two languages, such that words from either language that have similar meanings are close together~\cite{hermann2013multilingual,lauly2014autoencoder,gouws2014bilbowa}. For a test-of-concept experiment, we used English-French embeddings learned by the state-of-the-art BilBOWA model~\cite{gouws2014bilbowa} from the Wikipedia (monolingual) and Europarl (bilingual) corpora.\footnote{The approach should work with any bilingual embeddings. We thank Stephan Gouws for doing the training.} We trained the RNN model to map from English definitions to English words in the bilingual space. At test time, after reading an English definition, we then simply return the nearest French word neighbours to that definition.  

Because no benchmarks exist for quantitative evaluation of bilingual reverse dictionaries, we compare this approach qualitatively with two alternative methods for mapping definitions to words across languages. The first is analogous to the W2V Add model of the previous section: in the bilingual embedding space, we first compose the embeddings of the English words in the query definition with elementwise addition, and then return the French word whose embedding is nearest to this vector sum. The second uses the RNN monolingual reverse dictionary model to identify an English word from an English definition, and then translates that word using Google Translate.

Table~\ref{cross} shows that the RNN model can be effectively modified to create a cross-lingual reverse dictionary. It is perhaps unsurprising that the W2V Add model candidates are generally the lowest in quality given the performance of the method in the monolingual setting. In comparing the two RNN-based methods, the RNN (embedding space) model appears to have two advantages over the RNN + Google approach. First, it does not require online access to a bilingual word-word mapping as defined e.g. by Google Translate. Second, it less prone to errors caused by word sense ambiguity. For example, in response to the query \emph{an emotion you feel after being rejected}, the bilingual embedding RNN returns emotions or adjectives describing mental states. In contrast, the monolingual+Google model incorrectly maps the plausible English response \emph{regret} to the verbal infinitive \emph{regretter}. The model makes the same error when responding to a description of a fly, returning the verb \emph{voler} (to fly).

\subsection{Discussion}

We have shown that simply training RNN or BOW NLMs on six dictionaries yields a reverse dictionary that performs comparably to the leading commercial system, even with access to much less dictionary data. Indeed, the embedding models consistently return syntactically and semantically plausible responses, which are generally part of a more coherent and homogeneous set of candidates than those produced by the commercial systems. We also showed how the architecture can be easily extended to produce bilingual versions of the same model. 

In the analyses performed thus far, we only test the dictionary embedding approach on tasks that it was trained to accomplish
 (mapping definitions or descriptions to words). In the next section, we explore whether the knowledge learned by dictionary embedding models can be effectively transferred to a novel task. 

\section{General Knowledge (crossword) Question Answering}

The automatic answering of questions posed in natural language is a central problem of Artificial Intelligence. Although web search and IR techniques provide a means to find sites or documents related to language queries, at present, internet users requiring a specific fact must still sift through pages to locate the desired information. 

Systems that attempt to overcome this, via fully open-domain or general knowledge question-answering (open QA), generally require large teams of researchers, modular design and powerful infrastructure, exemplified by IBM's Watson~\cite{ferrucci2010building}. For this reason, much academic research focuses on settings in which the scope of the task is reduced. This has been achieved by restricting questions to a specific topic or domain~\cite{molla2007question}, allowing systems access to pre-specified passages of text from which the answer can be inferred~\cite{Iyyer:Boyd-Graber:Claudino:Socher:Daume-2014,weston2015towards}, or centering both questions and answers on a particular knowledge base~\cite{berant14paraphrasing,bordes2014question}. 

In what follows, we show that the dictionary embedding models introduced in the previous sections may form a useful component of an open QA system. Given the absence of a knowledge base or web-scale information in our architecture, we narrow the scope of the task by focusing on general knowledge crossword questions. General knowledge (non-cryptic, or quick) crosswords appear in national newspapers in many countries. Crossword question answering is more tractable than general open QA for two reasons. First, models know the length of the correct answer (in letters), reducing the search space. Second, some crossword questions mirror definitions, in that they refer to fundamental properties of concepts (\emph{a twelve-sided shape}) or request a category member (\emph{a city in Egypt}).\footnote{As our interest is in the language understanding, we do not address the question of fitting answers into a grid, which is the main concern of end-to-end automated crossword solvers~\cite{littman2002probabilistic}.} 

\subsection{Evaluation} 

General Knowledge crossword questions come in different styles and forms. We used the Eddie James crossword website to compile a bank of sentence-like general-knowledge questions.\footnote{\url{http://www.eddiejames.co.uk/}} Eddie James is one of the UK's leading crossword compilers, working for several national newspapers. Our { \bf long} question set consists of the first 150 questions (starting from puzzle \#1) from his general-knowledge crosswords, excluding clues of fewer than four words and those whose answer was not a single word (e.g. \emph{kingjames}).

To evaluate models on a different type of clue, we also compiled a set of {\bf short}er questions based on the Guardian Quick Crossword. Guardian questions still require general factual or linguistic knowledge, but are generally shorter and somewhat more cryptic than the longer Eddie James clues. We again formed a list of 150 questions, beginning on 1 January 2015 and excluding any questions with multiple-word answers. For clear contrast, we excluded those few questions of length greater than four words. Of these 150 clues, a subset of 30 were {\bf single-word} clues. All evaluation datasets are available online with the paper. 

As with the reverse dictionary experiments, candidates are extracted from models by inputting definitions and returning words corresponding to the closest embeddings in the target space. In this case, however, we only consider candidate words \emph{whose length matches the length specified in the clue}.

\begin{table}[ht]
{\small
\emph
\hfill{}
\begin{tabular}{r|cl}
\bf Test set & \bf Word & \bf Description \\
\hline

\hdashline
Long &   \emph{Baudelaire} & "French poet \\ 
 (150) & & and key figure \\ 
&& in the development \\ 
&& of Symbolism." \\
\hdashline 
\rule{0pt}{3ex} 

Short (120) &   \emph{satanist} & "devil devotee" \\

\hdashline
\rule{0pt}{3ex} 
Single-Word (30) &   \emph{guilt} & "culpability" \\
\end{tabular}
\caption{Examples of the different question types in the crossword question evaluation dataset.}
\label{tb:tablename}}
\end{table}

\begin{table*}[ht]
{\small
\hfill{}
\begin{tabular}{l|ccc|ccc|ccc|}
\multicolumn{1}{c}{} & \multicolumn{3}{c}{{\bf Question Type }}& \multicolumn{6}{|c|}{\emph{avg rank -accuracy@10/100 - rank variance} } \\
\multicolumn{10}{c}{} \\
\multicolumn{1}{c}{\textbf{}}&\multicolumn{3}{|c|}{\textbf{Long (150)}}& \multicolumn{3}{|c|}{\textbf{Short (120)}} & \multicolumn{3}{|c|}{\textbf{Single-Word (30)}}  \\
\hline
One Across & & .39 / && & \bf .68 / &&& .70 / &  \\
Crossword Maestro& & .27 /&& & .43 /&&& .73 / & \\
\hdashline
W2V add & 42  & .31/.63 & 92  & 11 & .50/.78 & 66 & \bf 2 & \bf .79/.90 & 45  \\
\hdashline
RNN cosine	& 15 &  .43/.69 & 108 & 22 & .39/.67 & 117 & 72 & .31/.52 & 187 \\
RNN w2v cosine &	4 & .61/.82 & 60 & \bf 7 & .56/.79 & 60 & 12 & .48/.72 & 116 \\
RNN ranking 	& 6 & .58/.84 & \bf 48 & 10 & .51/.73 & 57 & 12 & .48/.69 & 67 \\
RNN w2v ranking &	\bf 3 & .62/.80 & 61 & 8 & .57/.78 & 49 & 12 & .48/.69 & 114 \\
BOW cosine & 4 & .60/.82 & 54 & \bf 7 & .56/.78 & 51 & 12 & .45/.72 & 137 \\
BOW w2v cosine & 4 & .60/.83 & 56 & \bf 7  & .54/.80 & 48 & 3 & .59/.79 & 111 \\
BOW ranking	& 5 & \bf .62/.87 & 50 & 8 & .58/\bf .83 & 37 & 8 & .55/.79 & \bf 39 \\
BOW w2v ranking & 5 & .60/.86 & \bf 48 & 8 & .56/.83 & \bf 35 & 4 & .55/.83 & 43 \\
\end{tabular}}
\hfill{}
\caption{Performance of different models on crossword questions of different length. The two commercial systems are evaluated via their web interface so only accuracy@10 can be reported in those cases. }
\label{results2}
\label{tb:tablename}
\end{table*}

\subsection{Benchmarks and Comparisons}

As with the reverse dictionary experiments, we compare RNN and BOW NLMs with a
simple unsupervised baseline of elementwise addition of Word2Vec vectors in the
embedding space (we discard the ineffective \emph{W2V mult} baseline), again
restricting candidates to words of the pre-specified length. We also compare to
two bespoke online crossword-solving engines. The first, One Across
(\url{http://www.oneacross.com/}) is the candidate generation module of the
award-winning \emph{Proverb} crossword system~\cite{littman2002probabilistic}.
Proverb, which was produced by academic researchers, has featured in national
media such as New Scientist, and beaten expert humans in crossword solving
tournaments. The second comparison is with Crossword Maestro
(\url{http://www.crosswordmaestro.com/}), a commercial crossword solving system
that handles both cryptic and non-cryptic crossword clues (we focus only on the
non-cryptic setting), and has also been featured in national
media.\footnote{
    See e.g.
    \url{http://www.theguardian.com/crosswords/crossword-blog/2012/mar/08/crossword-blog-computers-crack-cryptic-clues}
}
We are unable to compare against a third well-known automatic crossword solver,
\emph{Dr Fill}~\cite{ginsberg2011dr}, because code for Dr Fill's
candidate-generation module is not readily available. As with the RNN and
baseline models, when evaluating existing systems we discard candidates whose
length does not match the length specified in the clue.  

Certain principles connect the design of the existing commercial systems and differentiate them from our approach. Unlike the NLMs, they each require query-time access to large databases containing common crossword clues, dictionary definitions, the frequency with which words typically appear as crossword solutions and other hand-engineered and task-specific components~\cite{littman2002probabilistic,ginsberg2011dr}. 

\begin{table*}[ht]
{\small
\emph
\hfill{}
\begin{tabular}{r|ccccc|}
\bf Input Description & \bf One Across& \bf Crossword Maestro & \bf BOW &  \bf RNN  \\
\hline

\rule{0pt}{3ex} 

  "Swiss mountain & 1:\emph{noted} 2:\emph{front} & 1:\emph{after} 2:\emph{favor} & 1:\emph{\bf Eiger} 2.\emph{Crags}   &  1:\emph{\bf Eiger} 2:\emph{Aosta}  \\
peak famed for its & 3:\emph{\bf Eiger} 4:\emph{crown} & 3:\emph{ahead} 4:\emph{along} &   3:\emph{Teton} 4:\emph{Cerro}  & 3:\emph{Cuneo} 4:\emph{Lecco}\\
 north face (5)" & 5:\emph{fount} &  5:\emph{being} &  5:\emph{Jebel} &  5:\emph{Tyrol} \\
\rule{0pt}{3ex} 
  "Old Testament & 1:\emph{\bf Joshua} 2:\emph{Exodus} &  1:\emph{devise} 2:\emph{Daniel}& 1:\emph{Isaiah} 2:\emph{Elijah}   &  1:\emph{\bf Joshua} 2:\emph{Isaiah}  \\
successor to & 3:\emph{Hebrew} 4:\emph{person} &   3:\emph{Haggai} 4:\emph{ Isaiah}  &3:\emph{\bf Joshua} 4:\emph{Elisha}  & 3:\emph{Gideon} 4:\emph{Elijah}\\
 Moses (6)" & 5:\emph{across} & 5:\emph{Joseph}&  5:\emph{Yahweh} &  5:\emph{Yahweh} \\	
\rule{0pt}{3ex} 
  "The former & 1:\emph{Holland} 2:\emph{general} &  1:\emph{Holland} 2:\emph{ancient} & 1:\emph{\bf Guilder} 2:\emph{Holland}   &  1:\emph{\bf 	Guilder} 2:\emph{Escudos}  \\
currency of the  & 3:\emph{Lesotho} &   3:\emph{earlier} 4:\emph{onetime}&   3:\emph{Drenthe} 4:\emph{Utrecht}  & 3:\emph{Pesetas} 4:\emph{Someren}\\
 Netherlands&  &5:\emph{qondam}&  5:\emph{Naarden} &  5:\emph{Florins} \\
 (7)"&  \\
\rule{0pt}{3ex} 
  "Arnold, 20th & 1:\emph{surrealism} &  1:\emph{disharmony}  & 1:\emph{\bf Schoenberg}   &  1:\emph{Mendelsohn} \\
Century composer &  2:\emph{laborparty}  &   2:\emph{dissonance} &  2:\emph{Christleib}  &  2:\emph{Williamson}  \\
pioneer of &  3:\emph{tonemusics}  &  3:\emph{bringabout} &  3:\emph{Stravinsky}  &  3:\emph{Huddleston}   \\
 atonality &4:\emph{introduced}  & 4:\emph{constitute} &4:\emph{Elderfield} & 4:\emph{Mandelbaum} \\
(10)"& 5:\emph{\bf Schoenberg} & 5:\emph{triggeroff} & 5:\emph{Mendelsohn} &  5:\emph{Zimmerman}\\

\end{tabular}}
\hfill{}
\caption{Responses from different models to example crossword clues. In each case the model output is filtered to exclude any candidates that are not of the same length as the correct answer. BOW and RNN models are trained without Word2Vec input embeddings and cosine loss.}
\label{egs}
\end{table*}

\subsection{Results}

The performance of models on the various question types is presented in Table~\ref{results2}. When evaluating the two commercial systems, One Across and Crossword Maestro, we have access to web interfaces that return up to approximately 100 candidates for each query, so can only reliably record membership of the top ten (accuracy@10).

On the long questions, we observe a clear advantage for all dictionary embedding models over the commercial systems and the simple unsupervised baseline. Here, the best performing NLM (RNN with Word2Vec input embeddings and ranking loss) ranks the correct answer third on average, and in the top-ten candidates over 60\% of the time.
	
As the questions get shorter, the advantage of the embedding models diminishes. Both the unsupervised baseline and One Across answer the short questions with comparable accuracy to the RNN and BOW models. One reason for this may be the difference in form and style between the shorter clues and the full definitions or encyclopedia sentences in the dictionary training data. As the length of the clue decreases, finding the answer often reduces to generating synonyms (\emph{culpability - guilt}), or category members (\emph{tall animal - giraffe}). The commercial systems can retrieve good candidates for such clues among their databases of entities, relationships and common crossword answers. Unsupervised Word2Vec representations are also  known to encode these sorts of relationships (even after elementwise addition for short sequences of words)~\cite{mikolov2013distributed}. This would also explain why the dictionary embedding models with pre-trained (Word2Vec) input embeddings outperfom those with learned embeddings, particularly for the shortest questions.

\subsection{Qualitative Analysis}

A better understanding of how the different models arrive at their answers can be gained from considering specific examples, as presented in Table~\ref{egs}. The first three examples show that, despite the apparently superficial nature of its training data (definitions and introductory sentences) embedding models can answer questions that require factual knowledge about people and places. Another notable characteristic of these model is the consistent semantic appropriateness of the candidate set. In the first case, the top five candidates are all mountains, valleys or places in the Alps; in the second, they are all biblical names. In the third, the RNN model retrieves currencies, in this case performing better than the BOW model, which retrieves entities of various type associated with the Netherlands. Generally speaking (as can be observed by the web demo), the `smoothness' or consistency in candidate generation of  the dictionary embedding models is greater than that of the commercial systems. Despite its simplicity, the unsupervised W2V addition method is at times also surprisingly effective, as shown by the fact that it returns \emph{Joshua} in its top candidates for the third query.

The final example in Table~\ref{egs} illustrates the surprising power of the BOW model. In the training data there is a single definition for the correct answer \emph{Schoenberg}: \emph{United States composer and musical theorist (born in Austria) who developed atonal composition}. The only word common to both the query and the definition is 'composer' (there is no tokenization that allows the BOW model to directly connect \emph{atonal} and \emph{atonality}). Nevertheless, the model is able to infer the necessary connections between the concepts in the query and the definition to return Schoenberg as the top candidate. 

Despite such cases, it remains an open question whether, with more diverse training data, the world knowledge required for full open QA (e.g. secondary facts about \emph{Schoenberg}, such as his family) could be encoded and retained as weights in a (larger) dynamic network, or whether it will be necessary to combine the RNN with an external memory that is less frequently (or never) updated. This latter approach has begun to achieve impressive results on certain QA and entailment tasks~\cite{bordes2014question,graves2014neural,weston2015towards}.

\section{Conclusion}

Dictionaries exist in many of the world's languages. We have shown how these lexical resources can constitute valuable data for training the latest neural language models to interpret and represent the meaning of phrases and sentences. While humans use the phrasal definitions in dictionaries to better understand the meaning of words, machines can use the words to better understand the phrases. We used two dictionary embedding architectures - a recurrent neural network architecture with a long-short-term memory, and a simpler linear bag-of-words model - to explicitly exploit this idea. 

On the reverse dictionary task that mirrors its training setting, NLMs that embed all known concepts in a continuous-valued vector space perform comparably to the best known commercial applications despite having access to many fewer definitions. Moreover, they generate smoother sets of candidates and require no linguistic pre-processing or task-specific engineering. We also showed how the description-to-word objective can be used to train models useful for other tasks. NLMs trained on the same data can answer general-knowledge crossword questions, and indeed outperform commercial systems on questions containing more than four words. While our QA experiments focused on crosswords, the results suggest that a similar embedding-based approach may ultimately lead to improved output from more general QA and dialog systems and information retrieval engines in general.  

We make all code, training data, evaluation sets and both of our linguistic tools publicly available online for future research. In particular, we propose the reverse dictionary task as a comparatively general-purpose and objective way of evaluating how well models compose lexical meaning into phrase or sentence representations (whether or not they involve training on definitions directly). 

In the next stage of this research, we will explore ways to enhance the NLMs described here, especially in the question-answering context. The models are currently not trained on any question-like language, and would conceivably improve on exposure to such linguistic forms. We would also like to understand better how BOW models can perform so well with no `awareness' of word order, and whether there are specific linguistic contexts in which models like RNNs or others with the power to encode word order are indeed necessary. Finally, we intend to explore ways to endow the model with richer world knowledge. This may require the integration of an external memory module, similar to the promising approaches proposed in several recent papers~\cite{graves2014neural,weston2015towards}.

\section*{Acknowledgments}

KC and YB acknowledge the support of the following organizations: NSERC, Calcul
Qu\'{e}bec, Compute Canada, the Canada Research Chairs and CIFAR. FH and AK
were supported by Google Faculty Research Award, and FH further by Google
European Doctoral Fellowship.


\bibliographystyle{acl}

\bibliography{iclr2015}

\end{document}